\documentclass[journal]{IEEEtran}

\ifCLASSINFOpdf
\else
\fi

\usepackage{xcolor}
\usepackage{graphicx}
\usepackage{algorithm}
\usepackage{algpseudocode}
\usepackage{amsmath}
\usepackage{amssymb}

\usepackage{longtable}
\usepackage{array}
\usepackage{geometry}
\usepackage{graphicx}
\usepackage{subcaption}

\begin{document}

\title{
Modeling Quantum Autoencoder Trainable Kernel for IoT Anomaly Detection}

\author{Swathi Chandrasekhar, Shiva Raj Pokhrel,  Swati Kumari and  Navneet Singh\thanks{Authors are from the Quantum Research Group at School of IT, Deakin University, Geelong, Australia; email: shiva.pokhrel@deakin.edu.au.}}
  \vspace{-15 mm}

\markboth{Journal of \LaTeX\ Class Files,~Vol.~14, No.~8, August~2025}%
{Shell \MakeLowercase{\textit{et al.}}: }

\maketitle
\begin{abstract}
Escalating cyber threats and the high-dimensional complexity of IoT traffic have outpaced classical anomaly detection methods. While deep learning offers improvements, computational bottlenecks limit real-time deployment at scale. We present a quantum autoencoder (QAE) framework that compresses network traffic into discriminative latent representations and employs quantum support vector classification (QSVC) for intrusion detection. Evaluated on three datasets, our approach achieves improved accuracy on ideal simulators and on the IBM Quantum hardware (ibm\_fez)—demonstrating practical quantum advantage on current NISQ devices. Crucially, moderate depolarizing noise acts as implicit regularization, stabilizing training and enhancing generalization. This work establishes quantum machine learning as a viable, hardware-ready solution for real-world cybersecurity challenges.

\end{abstract}


\IEEEpeerreviewmaketitle

\section{Introduction}

Securing Internet of Things (IoT) networks demands anomaly detectors~\cite{soe2019rule, pang2021deep, hdaib2024quantum} that can model high-dimensional, nonstationary traffic with limited compute budgets. Classical deep models help, but their footprint and training instability under distribution drift are problematic at the network edge.

The development of deep learning techniques brought unprecedented capabilities to anomaly detection~\cite{pang2021deep}. Neural networks, particularly autoencoders and recurrent neural networks (RNNs), demonstrated the ability to learn complex patterns in high-dimensional data without extensive feature engineering. Autoencoders, for instance, were used to compress network traffic data~\cite{javaid2016deep} into lower-dimensional representations and identify anomalies based on reconstruction errors~\cite{sewak2020overview}. These methods were further enhanced by distributed computing frameworks,which enabled the analysis of massive datasets. However, even with these improvements, deep learning models remain constrained by the computational limits of classical hardware, especially in real-time applications for large-scale networks.

\begin{figure*}[!t]
    \centering
    \includegraphics[width=\textwidth]{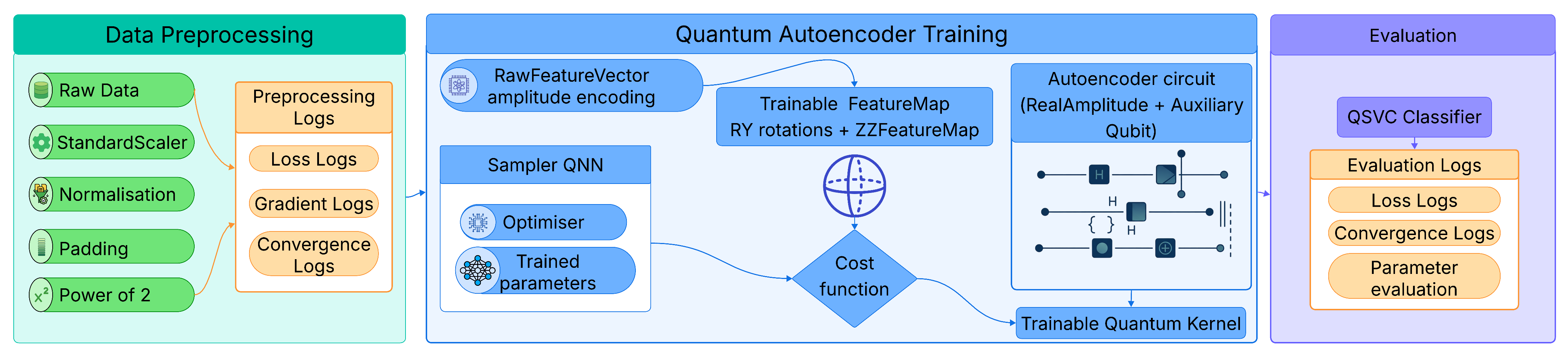}
    \caption{Overall pipeline of the proposed quantum autoencoder–based anomaly detection framework. (i) data preprocessing, including standardization, normalization, and power-of-two padding (ii) quantum autoencoder training, where RawFeatureVector amplitude encoding, a trainable feature map, and the autoencoder circuit are optimized via a sampler QNN and cost function to learn compressed latent representations and a trainable quantum kernel; and (iii) evaluation, in which a QSVC classifier uses the learned quantum kernel to perform anomaly detection.}
    \label{fig:architecture}
\end{figure*}

Quantum computing~\cite{oh2024quantum} has emerged as a promising solution to these challenges. By leveraging the principles of superposition, entanglement, and interference, quantum computers can process high-dimensional data and solve optimization problems exponentially faster than classical systems~\cite{oh2024quantum, singh2025modeling}. Quantum algorithms offer significant speedups for tasks like dimensionality reduction and clustering. In the context of network intrusion detection, quantum machine learning models, including quantum support vector machines (QSVMs) and quantum autoencoders (QAE), have the potential to detect complex and subtle anomalies~\cite{hdaib2024quantum}. In Fig.~\ref{fig:architecture}, we present a high level view of or idea of the end-to-end quantum autoencoder framework for anomaly detection that standardizes and encodes data, jointly learns compression and a quantum kernel, and then classifies anomalies using a QSVC in the optimized quantum feature space. 

%\subsection*{Key Contributions}
In this paper, we propose a quantum-enhanced anomaly detection framework that integrates coherence-driven feature compression, task-adaptive kernel learning, and noise-aware optimization. A SWAP-test–based QAE compresses high-dimensional traffic data into compact latent states while preserving discriminative structure, and a trainable quantum kernel—optimized jointly with the QAE via SPSA—aligns the latent space to the downstream classification task. We show that moderate depolarizing noise acts as an implicit regularizer that stabilizes gradients and improves generalization. The resulting QAE--QSVC model achieves over 92\% accuracy on Bot\_IoT, IoT23, and KDD99 in noiseless simulations, and maintains $\ge$83\% accuracy on IBM Quantum hardware, demonstrating practical feasibility. The design remains resource-efficient through amplitude encoding with $n_q=\lceil\log_2 d\rceil$ qubits and shallow circuits, enabling deployment in constrained IoT settings. Our pipeline (Fig.~\ref{fig:architecture}) performs classical preprocessing, amplitude encoding, QAE-based latent learning, task-adapted kernel optimization, and QSVC training, evaluated under both ideal and noisy quantum regimes.

\subsection{Data Encoding and Preprocessing}

The preprocessed dataset samples were subjected to a three-stage pipeline to ensure compatibility with quantum state representation. Each sample $\mathbf{x} \in \mathbb{R}^d$ underwent standardization via z-score normalization: $\mathbf{x}_{\text{scaled}} = \frac{\mathbf{x} - \mu}{\sigma}$, where $\mu$ and $\sigma$ represent the feature-wise mean and standard deviation computed across the training corpus. Subsequently, L2 normalization was applied to satisfy the unit norm constraint required for quantum amplitude encoding: $\mathbf{x}_{\text{norm}} = \frac{\mathbf{x}_{\text{scaled}}}{\|\mathbf{x}_{\text{scaled}}\|_2}$.

To map the $d$-dimensional feature space onto a quantum register, we computed the minimum qubit requirement as $n_q = \lceil\log_2(d)\rceil$, yielding a $2^{n_q}$-dimensional Hilbert space. Zero-padding was applied to extend each sample to match this dimension: $\mathbf{x}_{\text{padded}} = [\mathbf{x}_{\text{norm}}, \underbrace{0, \ldots, 0}_{2^{n_q}-d}]^T$.

\textbf{Amplitude Encoding Implementation}: The \texttt{RawFeatureVector}~\cite{sahin2025qiskitmachinelearningopensource} circuit directly encodes the classical feature vector into quantum amplitudes via state preparation:
% \begin{equation}
$|\psi(\mathbf{x})\rangle = \sum_{i=0}^{2^{n_q}-1} x_i |i\rangle$
% \end{equation}
where $x_i$ represents the $i$-th component of $\mathbf{x}_{\text{padded}}$. This encoding scheme, implemented through the feature map \texttt{fm = RawFeatureVector(2**num\_qubits)}, maps classical data into the computational basis states of an $n_q$-qubit system, preserving the L2 norm as the total probability: $\sum|x_i|^2 = 1$.

\subsection{Modeling Trainable Quantum Autoencoder Kernel}

The QAE was designed to compress the $n_q$-qubit encoded state into an $n_l$-qubit latent representation, discarding $n_t$ qubits of redundant information, where $n_t = n_q - n_l$. The architecture employs a $(n_l + 2n_t + 1)$-qubit system partitioned into a latent register of $n_l$ qubits preserving compressed information, trash register A of $n_t$ qubits containing discarded information, trash register B of $n_t$ reference qubits initialized to $|0\rangle$, and an auxiliary qubit for fidelity measurement~\cite{mirsky2018kitsune}.

\subsubsection{Encoder Circuit}

The variational encoder circuit $U_{\text{enc}}(\boldsymbol{\theta})$ was constructed using the \texttt{RealAmplitudes} ansatz with n repetitions, applied to the first $(n_l + n_t)$ qubits:
% \begin{equation}
$U_{\text{enc}}(\boldsymbol{\theta}) = \prod_{r=1}^{n} \left[ U_{\text{ent}} \cdot U_{\text{rot}}(\boldsymbol{\theta}_r) \right]$
% \end{equation}
where $U_{\text{rot}}$ consists of parameterized RY rotations:
% \begin{equation}
$U_{\text{rot}}(\boldsymbol{\theta}_r) = \bigotimes_{j=0}^{n_l + n_t - 1} RY(\theta_{r,j})$
% \end{equation}
and $U_{\text{ent}}$ implements a linear entanglement pattern using CX gates connecting adjacent qubits. The total parameter count for the encoder is $n \times (n_l + n_t)$ parameters.

\subsubsection{SWAP Test for Fidelity Measurement}

The compression quality was evaluated using a quantum SWAP test circuit that measures the fidelity between trash registers A and B. The auxiliary qubit was prepared in superposition, followed by controlled-SWAP operations.
% \begin{equation}
% |\psi_{\text{aux}}\rangle = H|0\rangle = \frac{1}{\sqrt{2}}(|0\rangle + |1\rangle)
% \end{equation}
% \begin{equation}
% \text{CSWAP}_{\text{aux}}(\text{trash A}_i, \text{trash B}_i) \quad \forall i \in \{0, \ldots, n_t-1\}
% \end{equation}
After applying the inverse Hadamard $H^\dagger$, the measurement outcome probability on the auxiliary qubit.
% is:
% \begin{equation}
% P(|1\rangle) = \frac{1}{2}\left(1 - |\langle\psi_{\text{trash A}}|\psi_{\text{trash B}}\rangle|^2\right)
% \end{equation}
The reconstruction loss $L_{\text{AE}}$ is defined as the expectation of measuring $|1\rangle$, which quantifies the information retained in the trash qubits
% \begin{equation}
% L_{\text{AE}}(\boldsymbol{\theta}) = \mathbb{E}_{\mathbf{x} \sim \mathcal{D}}[P(|1\rangle | \mathbf{x}, \boldsymbol{\theta})]
% \end{equation}
Optimal compression is achieved when trash A and trash B are maximally entangled with the latent space, yielding $P(|1\rangle) \rightarrow 0$.

\subsubsection{Training Protocol}

The autoencoder was trained using the COBYLA optimizer, a gradient-free derivative of Powell's method that constructs successive linear approximations of the objective function. The loss function was estimated via mini-batch sampling, where multiple batches were randomly drawn from the training pool at each iteration. The stochastic batching strategy provides variance estimates for loss uncertainty quantification:
% \begin{equation}
$\sigma_{\text{loss}} = \sqrt{\frac{1}{B-1}\sum_{b=1}^{B}(\hat{L}_b - \hat{L}_{\text{AE}})^2}$
% \end{equation}
where $B$ denotes the number of batches per iteration and $\hat{L}_b$ represents the loss computed on batch $b$. This variance was logged alongside 95\% confidence intervals: $[\mu - 1.96\sigma, \mu + 1.96\sigma]$.

\subsubsection{Latent Feature Extraction}

Post-training, the encoder circuit (excluding the SWAP test) was used to extract compressed representations. For each input $\mathbf{x}_{\text{padded}}$, the quantum state after encoding is:
% \begin{equation}
$|\phi(\mathbf{x}, \boldsymbol{\theta}^*)\rangle = U_{\text{enc}}(\boldsymbol{\theta}^*) |\psi(\mathbf{x})\rangle$
% \end{equation}
where $\boldsymbol{\theta}^*$ denotes the optimized parameters. The latent features were extracted by computing the probability distribution over the $n_l$-qubit latent subspace.
% \begin{equation}
% \mathbf{z} = [p_0, p_1, \ldots, p_{2^{n_l}-1}]^T, \quad p_i = |\langle i | \phi(\mathbf{x}, \boldsymbol{\theta}^*) \rangle|^2
% \end{equation}
% where $|i\rangle$ represents computational basis states of the latent register ($i \in \{0, \ldots, 2^{n_l}-1\}$). This transformation maps the original $d$-dimensional feature space to a $2^{n_l}$-dimensional quantum feature space: $\mathbb{R}^d \rightarrow \mathbb{C}^{2^{n_q}} \rightarrow \mathbb{R}^{2^{n_l}}$.

% The information content of each latent representation was quantified using von Neumann entropy:
% \begin{equation}
% S(\mathbf{z}) = -\sum_{i=0}^{2^{n_l}-1} p_i \log_2(p_i)
% \end{equation}

\subsection{Trainable Quantum Kernel}

% \subsubsection{Kernel Architecture}

% The quantum kernel function was constructed using a hybrid trainable feature map combining parameterized single-qubit rotations with data-dependent entangling operations. Given the $2^{n_l}$-dimensional latent features $\mathbf{z}$, we employed an $n_k$-qubit feature map consisting of two components:

% \textbf{Trainable Rotation Layer}:
% \begin{equation}
% U_{\text{train}}(\boldsymbol{\phi}) = \bigotimes_{j=0}^{n_k-1} RY(\phi_j)
% \end{equation}
% where $\boldsymbol{\phi} = [\phi_0, \phi_1, \ldots, \phi_{n_k-1}]$ are trainable parameters.

% \textbf{Data Encoding Layer}: The \texttt{ZZFeatureMap} with multiple repetitions implements alternating single-qubit Z-rotations and two-qubit ZZ-entangling gates. The single-qubit layer applies $U_Z(\mathbf{z}) = \bigotimes_{j=0}^{n_k-1} e^{i z_j Z_j}$, while the entangling layer creates pairwise interactions through $e^{i (\pi - z_j)(\pi - z_{j+1}) Z_j Z_{j+1}}$ for adjacent qubit pairs. The complete feature map is:
% \begin{equation}
% \Phi(\mathbf{z}, \boldsymbol{\phi}) = U_{\Phi}(\mathbf{z}) \cdot U_{\text{train}}(\boldsymbol{\phi})
% \end{equation}

%\subsubsection{Quantum Kernel Function}

The kernel matrix element between samples $\mathbf{z}_i$ and $\mathbf{z}_j$ is computed via the ComputeUncompute fidelity protocol:
\begin{equation}
K(\mathbf{z}_i, \mathbf{z}_j; \boldsymbol{\phi}) = \left|\langle 0^{\otimes n_k} | \Phi^\dagger(\mathbf{z}_j, \boldsymbol{\phi}) \Phi(\mathbf{z}_i, \boldsymbol{\phi}) | 0^{\otimes n_k} \rangle\right|^2
\end{equation}
This is estimated via sampling with a fixed number of shots per circuit execution
% \begin{equation}
$\hat{K}(\mathbf{z}_i, \mathbf{z}_j; \boldsymbol{\phi}) = \frac{N_{|0^{\otimes n_k}\rangle}}{N_{\text{shots}}}$
% \end{equation}
where $N_{|0^{\otimes n_k}\rangle}$ is the count of all-zero measurement outcomes.

%\subsubsection{Kernel Training Objective}

The kernel parameters $\boldsymbol{\phi}$ were optimized using the QSVC loss function, which maximizes the margin in the quantum feature space~\cite{Chandrasekhar2025}. For a binary classification task with labels $y \in \{-1, +1\}$, the loss is
% \begin{equation}
$L_{\text{QSVC}}(\boldsymbol{\phi}) = \frac{1}{2}\boldsymbol{\alpha}^T \mathbf{K}(\boldsymbol{\phi}) \boldsymbol{\alpha} - \sum_{i=1}^{N} \alpha_i$
% \end{equation}
% subject to:
% \begin{equation}
% \sum_{i=1}^{N} \alpha_i y_i = 0, \quad 0 \leq \alpha_i \leq C
% \end{equation}
% where $\boldsymbol{\alpha}$ are the dual variables obtained by solving the kernel SVM optimization, and $\mathbf{K}(\boldsymbol{\phi})$ is the kernel Gram matrix with elements $K_{ij} = y_i y_j K(\mathbf{z}_i, \mathbf{z}_j; \boldsymbol{\phi})$.

The kernel parameters were optimized using the SPSA algorithm. SPSA is a gradient-free stochastic optimization method particularly suited for noisy objective functions, as it estimates the gradient using only two function evaluations per iteration regardless of the parameter dimension. The hyperparameters include maximum iterations, learning rate $\alpha_k$, and perturbation magnitude $c_k$. The gradient estimate at iteration $k$ is computed via symmetric finite differences:
% \begin{equation}
$\hat{\nabla} L(\boldsymbol{\phi}_k) = \frac{L(\boldsymbol{\phi}_k + c_k \boldsymbol{\Delta}_k) - L(\boldsymbol{\phi}_k - c_k \boldsymbol{\Delta}_k)}{2c_k} \boldsymbol{\Delta}_k$
% \end{equation}
where $\boldsymbol{\Delta}_k$ is a random perturbation vector with elements sampled from $\{-1, +1\}$. 
% The parameter update rule is:
% \begin{equation}
% \boldsymbol{\phi}_{k+1} = \boldsymbol{\phi}_k - \alpha_k \hat{\nabla} L(\boldsymbol{\phi}_k)
% \end{equation}

\subsection{Quantum Support Vector Classification}

The optimized quantum kernel $K(\mathbf{z}_i, \mathbf{z}_j; \boldsymbol{\phi}^*)$ was used to train a QSVC by solving the optimization problem:
\begin{equation}
\max_{\boldsymbol{\alpha}} \sum_{i=1}^{N} \alpha_i - \frac{1}{2}\sum_{i=1}^{N}\sum_{j=1}^{N} \alpha_i \alpha_j y_i y_j K(\mathbf{z}_i, \mathbf{z}_j; \boldsymbol{\phi}^*)
\end{equation}
subject to $\sum_{i=1}^{N} \alpha_i y_i = 0$ and $0 \leq \alpha_i \leq C$, where $C$ is the regularization parameter that controls the trade-off between the maximization of the margin and the minimization of the training error. 
The QSVC exploits the capacity of the quantum kernel to implicitly embed data in an exponentially large feature space, where the kernel function $K(\mathbf{z}_i,\mathbf{z}_j;\boldsymbol{\phi}^*)$ corresponds to the inner product of quantum states. The entanglement structure encoded in the feature map captures complex non-linear relationships that are difficult to model classically, while the support vectors and their dual coefficients $\alpha_i$ define the decision boundary in this high-dimensional quantum space.

\section{Results}

\begin{table}[h!]
\centering
\begin{tabular}{|l|c|c|c|}
\hline
Dataset & Avg $loss\_std$ & Avg $lower\_bound$ & Avg $upper\_bound$ \\ 
\hline
Bot\_IoT & 0.0191 & 0.1843 & 0.2590 \\ 
IoT23  & 0.0171 & 0.2576 & 0.3244 \\ 
KDD & 0.0213 & 0.2826 & 0.3660 \\ 
\hline
\multicolumn{4}{|c|}{Depolarizing Noise}\\
\hline
Bot\_IoT & 0.0168 & 0.2906 & 0.3564 \\ 
IoT23 & 0.0302 & 0.1513 & 0.2699 \\ 
KDD & 0.0169 & 0.2644 & 0.3308 \\ 

\hline
\end{tabular}
\caption{Average training loss variability and confidence bounds across datasets Bot\_IoT, IoT23, and KDD for the quantum autoencoder under both noiseless and depolarizing noise conditions.}
\label{tab:loss_summary}
\end{table}
\section{Performance Evaluation}
This section presents the experimental results obtained using the proposed QAE-QSVC for anomaly detection across multiple network intrusion datasets, Bot\_IoT, IoT23, and KDD99. Both ideal and noisy quantum scenarios were evaluated using Aer simulation backends and real IBM Quantum hardware (ibm\_fez).
\begin{figure*}[htbp]
    \centering
    \begin{subfigure}[b]{0.32\textwidth}
        \includegraphics[width=\linewidth]{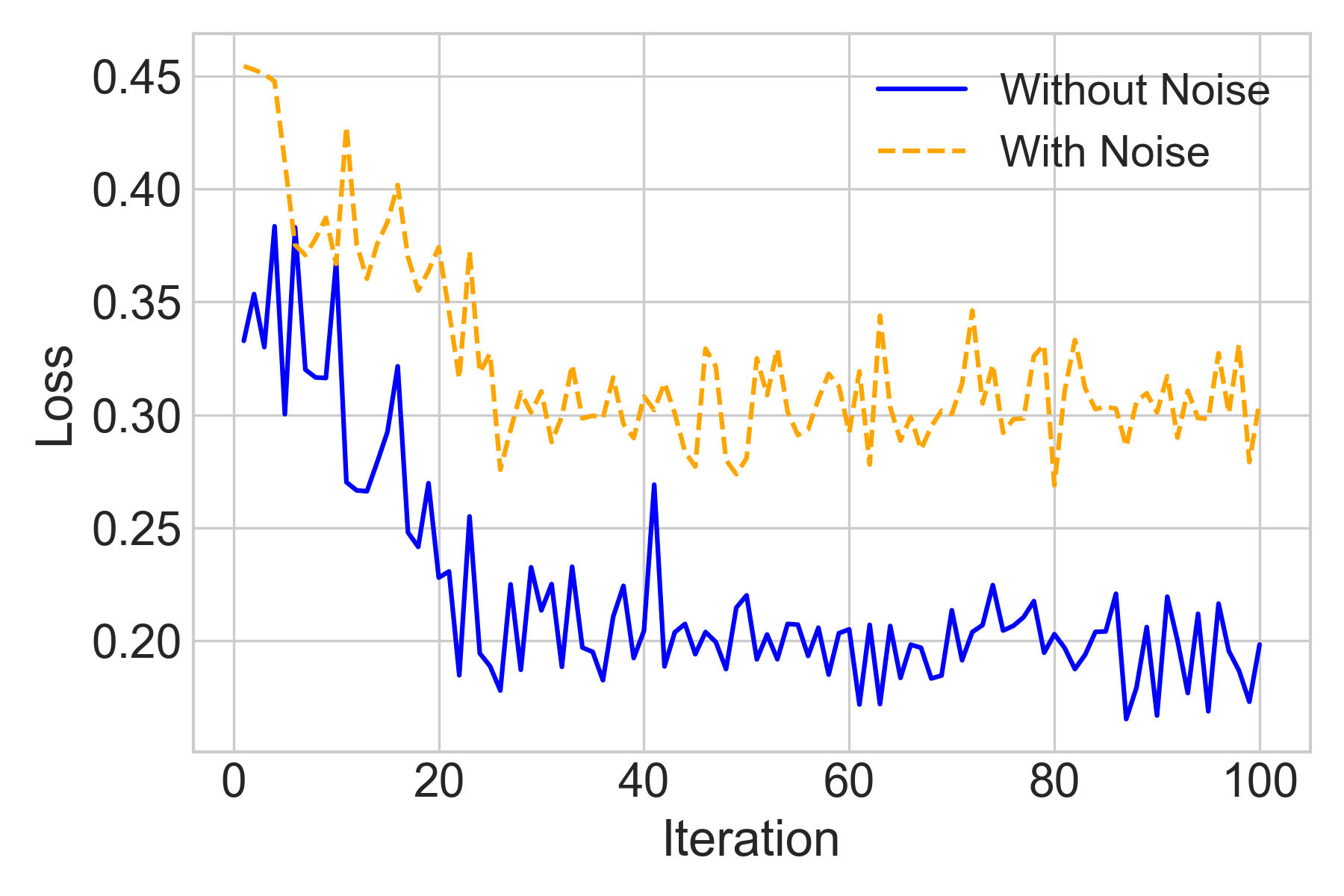}
        \caption{Bot\_IoT}
    \end{subfigure}
    \hfill
    \begin{subfigure}[b]{0.32\textwidth}
        \includegraphics[width=\linewidth]{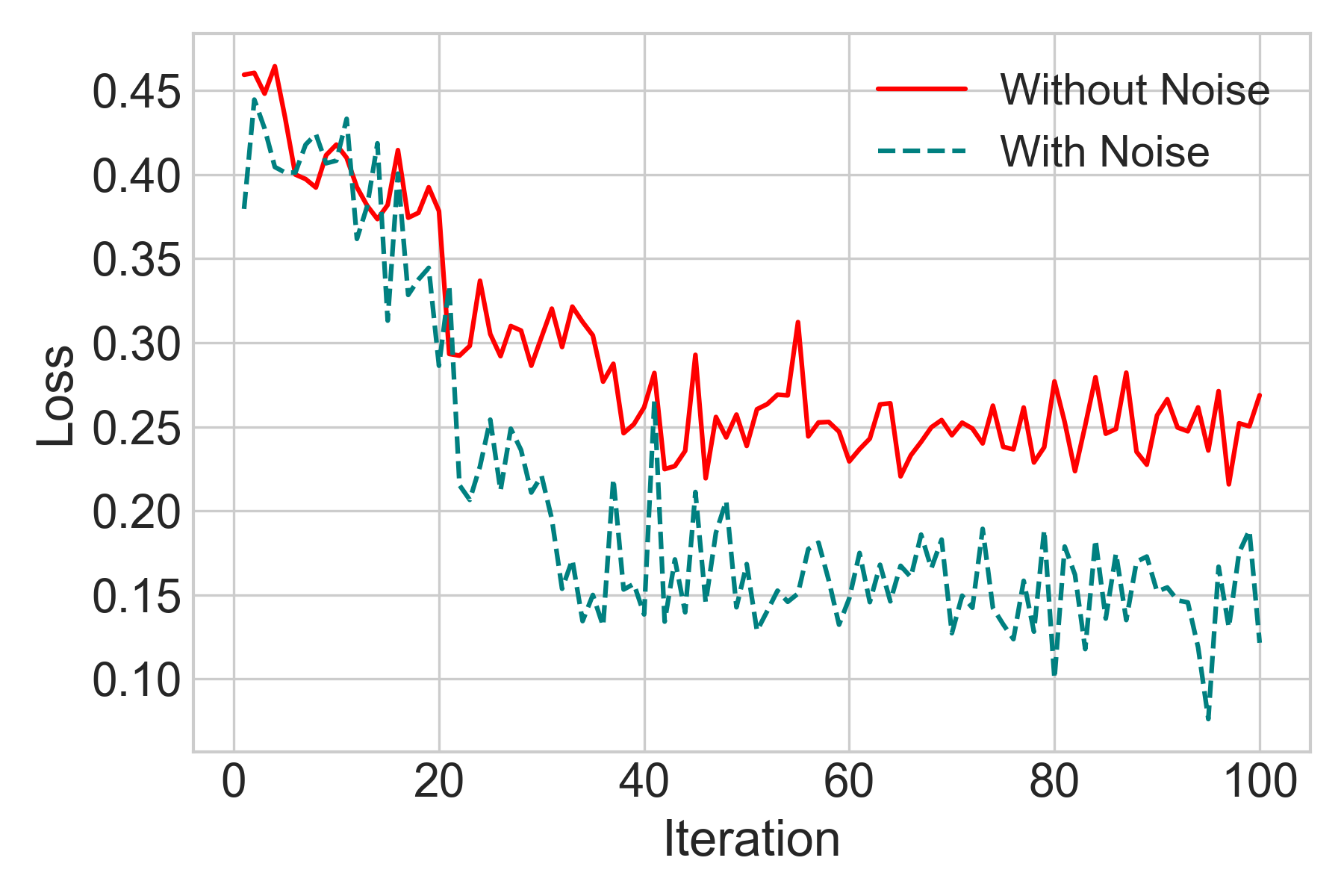}
        \caption{IoT23}
    \end{subfigure}
    \hfill
    \begin{subfigure}[b]{0.32\textwidth}
        \includegraphics[width=\linewidth]{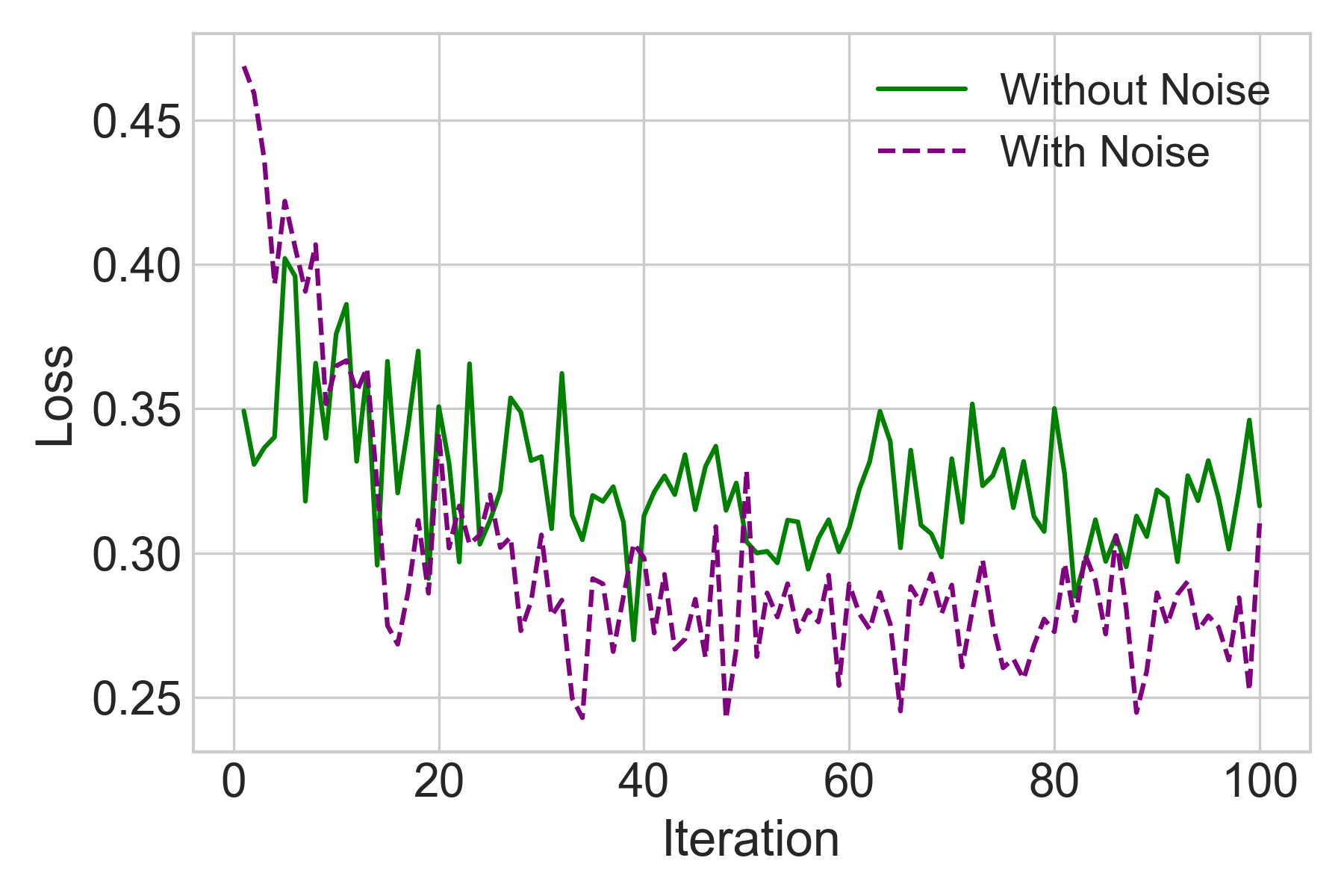}
        \caption{KDD}
    \end{subfigure}
    \caption{Training loss curves for the quantum autoencoder on Bot\_IoT, IoT23, and KDD datasets under noiseless and depolarizing noise conditions showing convergence of model acreoss iteraions.}
    \label{fig:loss_curves}
\end{figure*}

\textit{Training Loss Analysis}.
Figure~\ref{fig:loss_curves} illustrates the convergence patterns for each dataset under noisy and noise-free conditions. As observed, the QAE demonstrated stable convergence, with loss values consistently decreasing across iterations. In the Bot\_IoT dataset, noise-free training yielded a lower loss floor ($\approx 0.20$) compared to the noisy configuration ($\approx 0.30$), indicating the sensitivity of the circuit to accumulated decoherence. Conversely, in IoT23 and KDD datasets, the inclusion of noise led to enhanced regularization, stabilizing training oscillations after initial iterations. Table~\ref{tab:loss_summary} summarizes the average standard deviation and bound metrics for each case, where the lower standard deviation across noise conditions reflects the robustness of our hybrid optimization loop.

\begin{figure}[h!]
    \centering
    \includegraphics[width=0.48\textwidth]{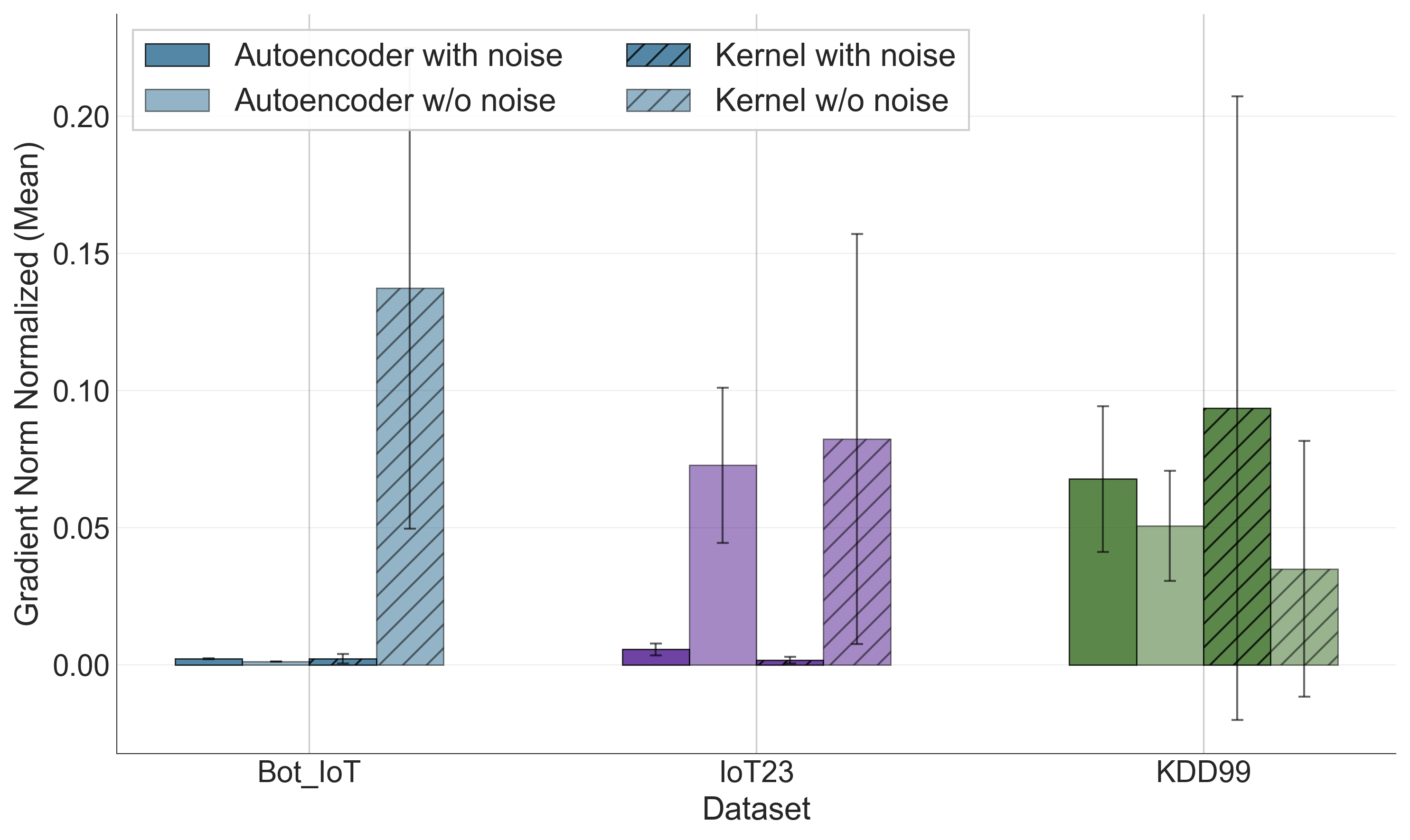}
    \caption{Normalized gradient for autoencoder and kernel training phases under depolarizing noise as well for Bot\_IoT, IoT23, and KDD}
    \label{fig:gradients}
\end{figure}

\textit{Gradient Behavior and Stability}.
Gradient norm analysis across all datasets Figure~\ref{fig:gradients} reveals distinct quantum noise propagation patterns. Bot\_IoT and KDD99 exhibited higher gradient magnitudes in the presence of noise, while IoT23 maintained comparatively smoother gradient transitions, demonstrating greater resilience of its latent representation. Normalized gradients emphasize reduced quantum circuit instability due to controlled learning rate and encoder compression.

\begin{figure}
    \centering
    \begin{subfigure}[b]{0.48\columnwidth}
        \includegraphics[width=\linewidth]{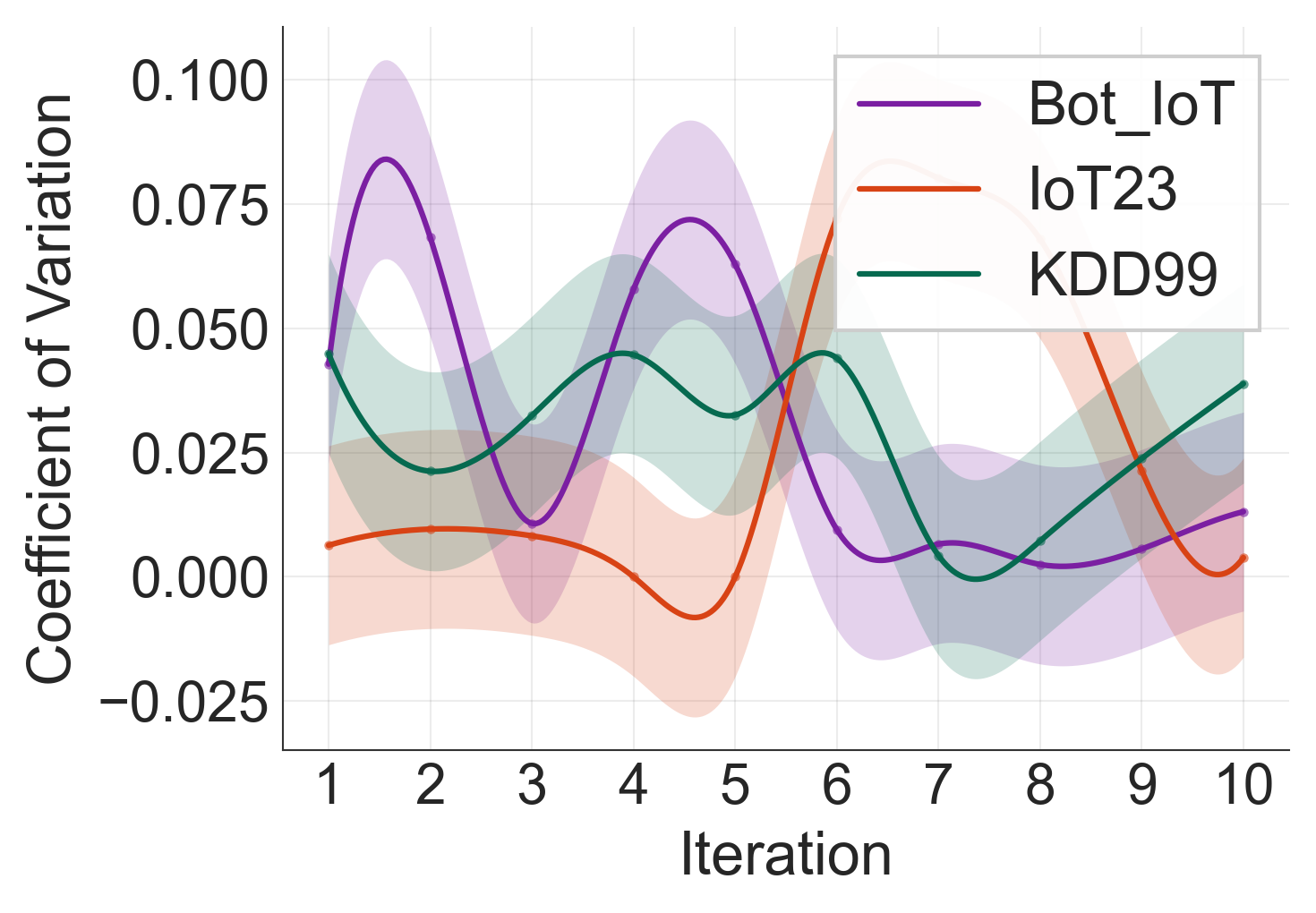}
        \caption{QSVC autoencoder variation with noise}
    \end{subfigure}
    \hfill
    \begin{subfigure}[b]{0.48\columnwidth}
        \includegraphics[width=\linewidth]{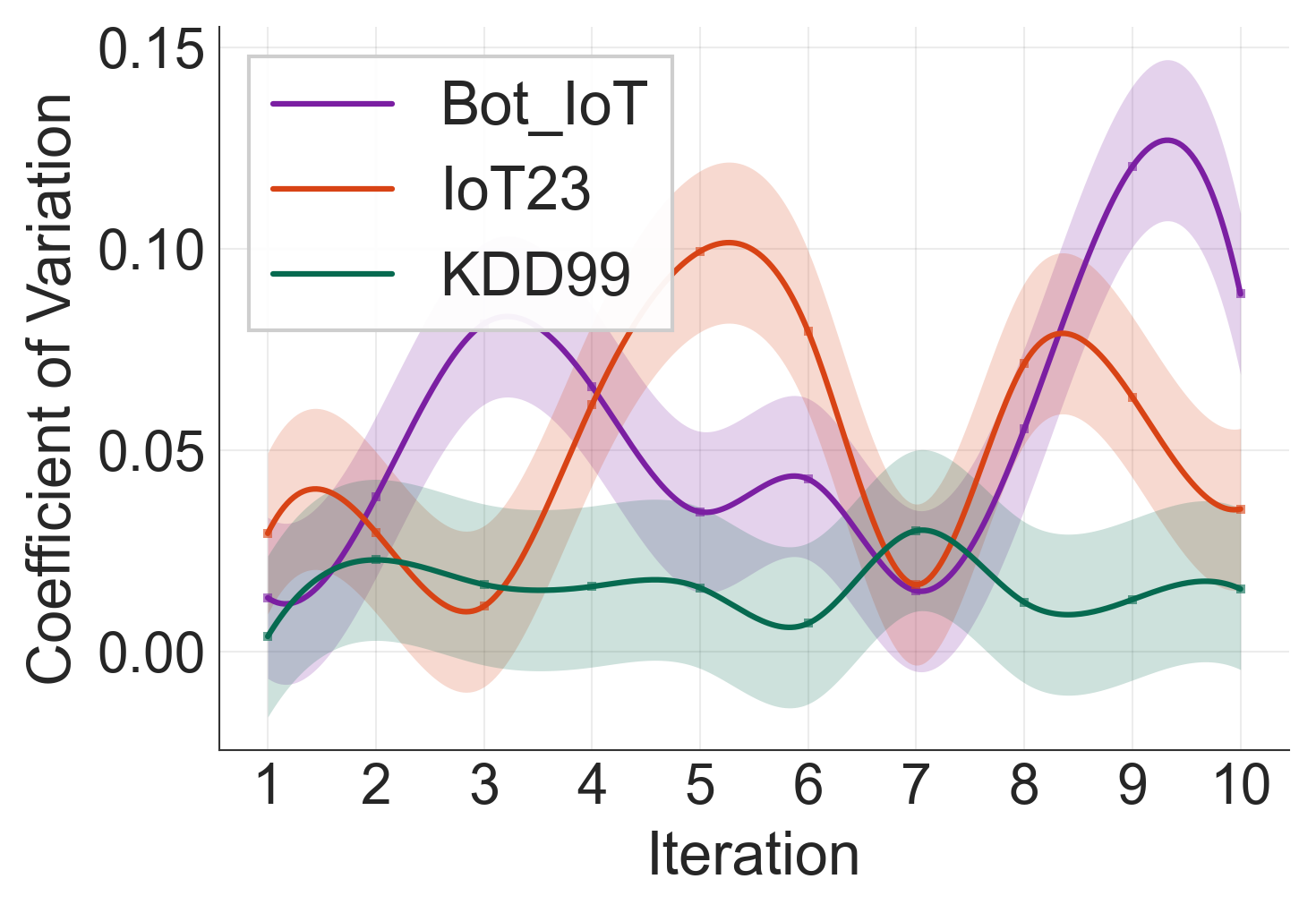}
        \caption{QSVC autoencoder variation without noise}
    \end{subfigure}
    
    \vspace{0.5em}
    
    \begin{subfigure}[b]{0.48\columnwidth}
        \includegraphics[width=\linewidth]{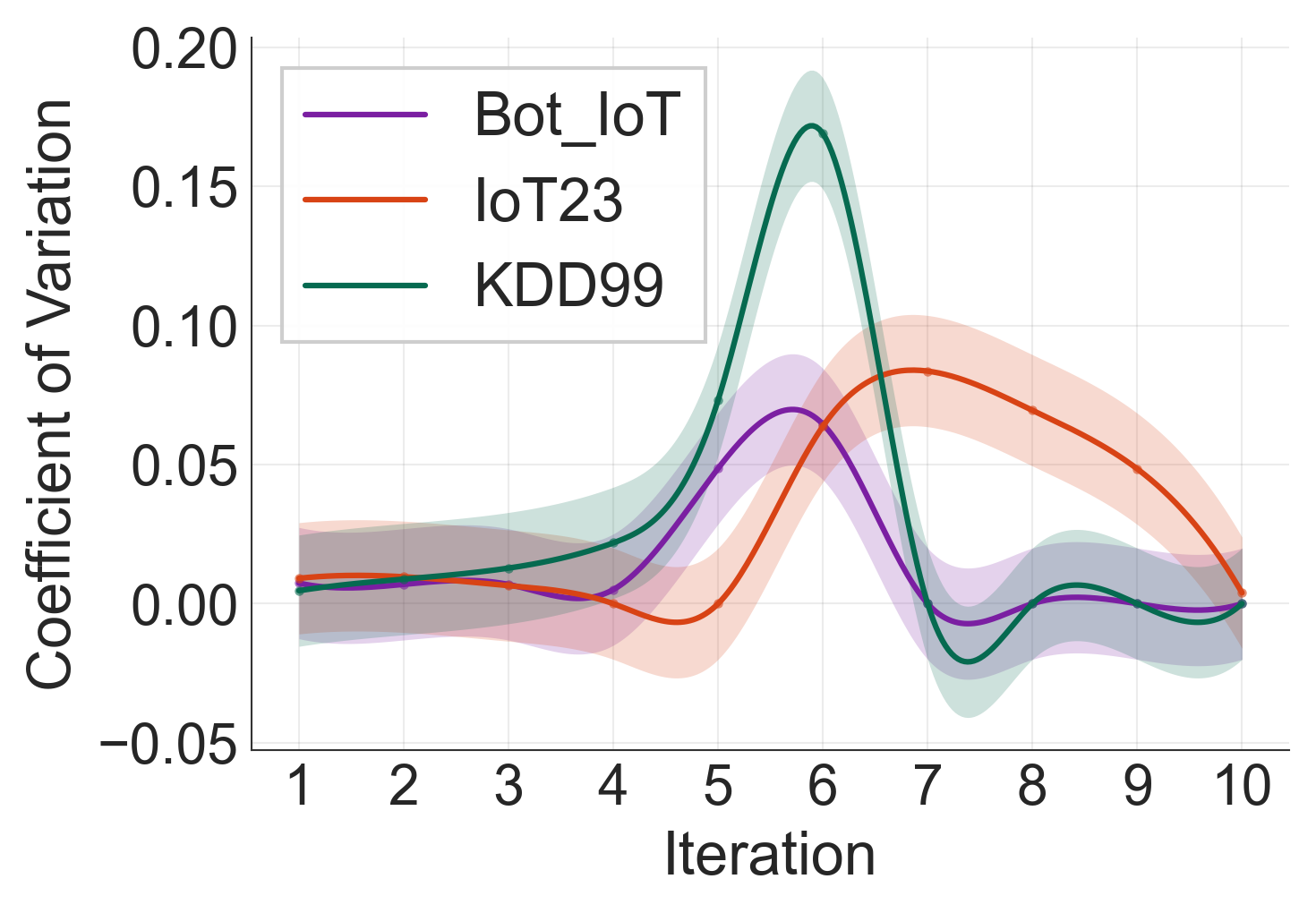}
        \caption{QSVC kernel variation with noise}
    \end{subfigure}
    \hfill
    \begin{subfigure}[b]{0.48\columnwidth}
        \includegraphics[width=\linewidth]{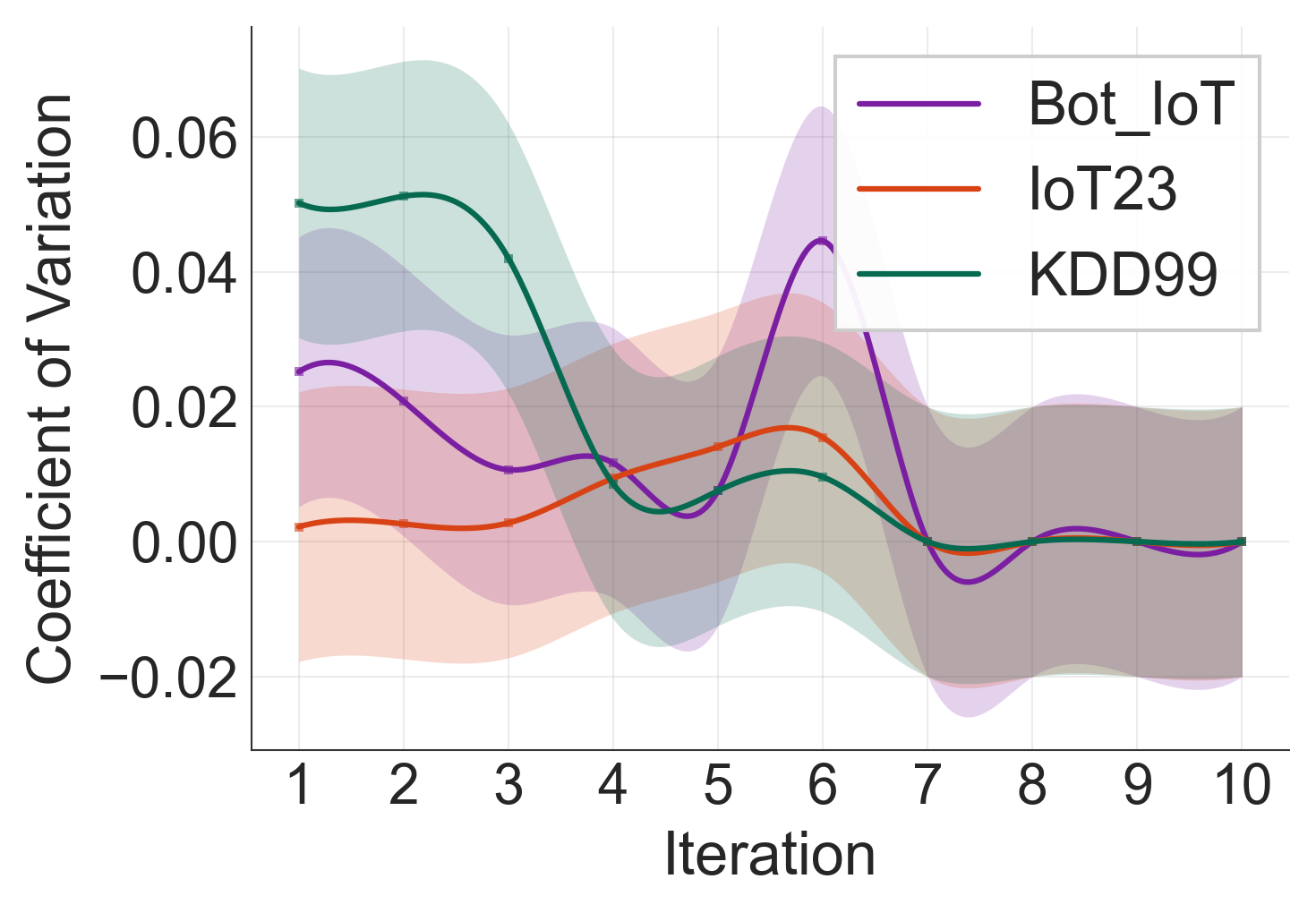}
        \caption{QSVC kernel variation without noise}
    \end{subfigure}
    
    \caption{Coefficient of variation of loss estimates for the QSVC autoencoder and kernel phases, comparing noisy and noiseless configurations.(a) and (b) show the variability during the autoencoder phase, while (c) and (d) depict the kernel phase across training iterations.}
    \label{fig:coeff_variation}
\end{figure}

\textit{Coefficient of Variation Analysis}.
Figures~\ref{fig:coeff_variation} present the coefficient of variation computed for autoencoder reconstruction and kernel evaluations. Lower variation trends in most noisy cases demonstrate that depolarizing noise introduces slight regularization rather than destructive interference, stabilizing the feature space embedding, particularly evident in KDD99.

% \begin{figure}[h!]
%     \centering
%     \subfloat[Autoencoder layer output variation]{\includegraphics[width=0.48\textwidth]{coeff_of_variation_autoencoder.png}}
%     \subfloat[QSVC kernel variation]{\includegraphics[width=0.48\textwidth]{coeff_of_variation_kernel.png}}
%     \caption{Coefficient of variation across iterations.}
%     \label{fig:coeff_variation}
% \end{figure}
\begin{table*}[h!]
\centering
\begin{tabular}{|l|c|c|c|c|c|c|c|c|c|}
\hline
Dataset & Train Acc & Test Acc & Train Prec & Test Prec & Train Rec & Test Rec & Train F1 & Test F1 & Time (s) \\ 
\hline
Bot\_IoT & 0.9962 & 0.9242 & 0.9960 & 0.9247 & 0.9962 & 0.9242 & 0.9961 & 0.9294 & 157369.7\\
IoT23 & 0.9875 & 0.9950 & 0.9562 & 0.9650 & 0.9875 & 0.9500 & 0.9453 & 0.9268 & 99167.8\\
KDD99 & 0.9875 & 0.9766 & 0.9874 & 0.9543 & 0.9875 & 0.9857 & 0.9874 & 0.9766 & 84069.4\\
\hline
\multicolumn{10}{|c|}{Depolarizing Noise}\\
\hline
Bot\_IoT & 0.9734 & 0.9545 & 0.9748 & 0.9581 & 0.9738 & 0.9545 & 0.9734 & 0.9543 & 91706.6\\
IoT23 & 0.9833 & 0.9737 & 0.9973 & 0.9836 & 0.9833 & 0.9833 & 0.9283 & 0.9222 & 86906.5\\
KDD99 & 0.9875 & 0.9833 & 0.9884 & 0.9847 & 0.9874 & 0.9833 & 0.9876 & 0.9836 & 86245.4\\
\hline
\end{tabular}
\caption{Performance on of the QAE–QSVC architecture on the Aer simulator under noiseless and depolarizing noise settings.}
\label{tab:sim_results}
\end{table*}

\textit{Performance on Simulator and Real Hardware}.
Table~\ref{tab:sim_results} summarizes the performance achieved using the Aer simulator. Across all datasets, the QAE+QSVC achieved test accuracies above 92\%, confirming strong generalization after quantum state compression. Even with depolarizing noise enabled, the model retained accuracy above 95\% for most datasets.

\begin{figure}[h!]
    \centering
    \includegraphics[width=0.35\textwidth, height=0.20\textheight]{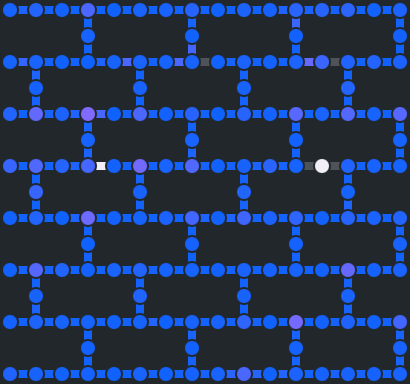}
    \caption{Performance of the QAE–QSVC model on IBM Quantum hardware \textit{(ibm\_fez)} for Bot\_IoT and IoT23, demonstrating the practical feasibility of deploying quantum autoencoder-based anomaly detection on current NISQ devices.}
    \label{fig:ibm_fez_map}
\end{figure}

Execution on the IBM Quantum system (\textit{ibm\_fez}), which consists of the qubit topology shown in Figure~\ref{fig:ibm_fez_map}, validated the feasibility of performing quantum autoencoder operations on real hardware despite calibration and readout errors. Due to runtime constraints (10 minutes) and limited qubit allocation, the tests were executed with reduced samples on two datasets. Even under such limitations, accuracies above 83\% were achieved, as shown in Table~\ref{tab:ibm_results}. These results strongly suggest that, with extended runtime and noise mitigation resources, the proposed QAE-QSVC approach could scale toward near-ideal accuracy.

\begin{table}[h!]
\centering
\begin{tabular}{|l|c|c|c|c|c|}
\hline
Dataset & Accuracy & Precision & Recall & F1-score & Time\\
\hline
Bot\_IoT & 0.8333 & 0.8809 & 0.8333 & 0.8333 & 353\\
IoT23 & 0.8437 & 0.7500 & 0.7500 & 0.7447 & 212\\
\hline
\end{tabular}
\caption{Qubit coupling map of the IBM Quantum \textit{ibm\_fez} device used for hardware experiments.}
\label{tab:ibm_results}
\end{table}

\noindent Overall, these results confirm that the proposed hybrid quantum framework not only achieves high classification performance under simulation but also sustains practical accuracy on real quantum devices. The consistent gradient stability, bounded loss, and low coefficient of variation reinforce the reliability and scalability of QAE based network anomaly detection.

% \subsection{Benchmarking against Hdaib et al.}

% To contextualize our results, we compare directly against the QAE based anomaly detection framework of~\cite{hdaib2024quantum}, specifically their Framework 1 (QAE + one-class SVM), on overlapping datasets. On \textbf{KDD99}, Hdaib et al. report 97.48\% accuracy and 97.19\% F1-score, whereas our QAE--QSVC achieves 98.33\% accuracy and 98.36\% F1 under depolarizing noise, an improvement of $\approx$1.2 percentage points. The performance gap is far more pronounced on \textbf{IoT23}: their Framework 1 yields only 82.53\% accuracy and 79.69\% F1, while our approach attains 97.37\% accuracy and 92.22\% F1 even with noise, corresponding to absolute gains of $\approx$15 and $\approx$13 percentage points, respectively (Table~\ref{tab:benchmark}).

% A critical methodological distinction is that~\cite{hdaib2023quantum} applied PCA before the quantum autoencoder, which discards variance and can remove discriminative structure. In contrast, we perform no classical dimensionality reduction; compression occurs solely within the coherence driven QAE, preserving task relevant information. This design choice, combined with supervised margin-based classification, explains the consistently superior detection performance of our QAE--QSVC framework.
To contextualize our results, we directly compare with the QAE-based anomaly detection framework of~\cite{hdaib2024quantum} (Framework~1: QAE + one-class SVM) on overlapping datasets. On \textit{KDD99}, their reported 97.48\% accuracy and 97.19\% F1 are surpassed by our noisy QAE--QSVC, which attains 98.33\% accuracy and 98.36\% F1. The gap is even larger on \textit{IoT23}: their 82.53\% accuracy and 79.69\% F1 contrast with our 97.37\% accuracy and 92.22\% F1 under depolarizing noise, corresponding to absolute gains of approximately 15 and 13 percentage points, respectively (Table~\ref{tab:benchmark}). The underlying difference is that~\cite{hdaib2023quantum} applies PCA before the QAE, discarding potentially discriminative variance. In sharp contrast, our approach performs no classical dimensionality reduction; all compression occurs within the coherence-driven QAE and is paired with supervised margin-based classification, yielding consistently superior detection performance.

\begin{table}[h]
\centering
\caption{Comparative evaluation of the proposed QAE–QSVC framework with ~\cite{hdaib2024quantum} Framework 1 on ideal and DP:depolarising noise}
\label{tab:benchmark}
\begin{tabular}{llcc}
\hline
\textbf{Dataset} & \textbf{Method} & \textbf{Test Acc} & \textbf{Test F1} \\
\hline
KDD99 & QAE + OC-SVM~\cite{hdaib2024quantum} & 97.48 & 97.19 \\
This work & QAE--QSVC (ideal) & 97.66 & 97.66 \\
    This work    & QAE--QSVC (DP) & \textbf{98.33} & \textbf{98.36} \\
\hline
IoT23 & QAE + OC-SVM~\cite{hdaib2024quantum} & 82.53 & 79.69 \\
   This work & QAE--QSVC (ideal) & \textbf{99.50} & \textbf{92.68 }\\
   This work & QAE--QSVC (DP) & 97.37 & 92.22 \\
\hline
\end{tabular}
\end{table}

\section{Conclusion}
This work shows that quantum autoencoding is a strong backbone for anomaly detection in large-scale IoT networks. By combining a coherence-driven Quantum Autoencoder (QAE) with a Quantum Support Vector Classifier (QSVC), we build a hybrid model that learns compact, discriminative latent representations directly from amplitude-encoded traffic data. The approach achieves high detection accuracy across multiple intrusion datasets while requiring far fewer computational resources than classical deep learning methods. A SWAP-test--based fidelity objective preserves task-relevant structure, and a trainable quantum kernel improves separability in the latent space. Experiments under ideal and noisy conditions demonstrate robustness, with quantum noise acting as a useful regularizer that enhances generalization. Tests on real IBM Quantum hardware further confirm the framework’s practicality. %Overall, the results position quantum autoencoders as a promising, resource-efficient direction for security analytics in emerging quantum–classical systems, with future work focused on scaling to larger qubit devices, improved noise mitigation, and hardware-aware ansatz design.

% This work demonstrates that quantum autoencoding can serve as an effective backbone for anomaly detection in large-scale IoT networks. By integrating a coherence-driven Quantum Autoencoder (QAE) with a Quantum Support Vector Classifier (QSVC), we present a hybrid architecture capable of learning compact, discriminative latent manifolds directly from amplitude-encoded network traffic. The resulting model achieves competitive detection accuracy across multiple intrusion datasets while operating on significantly fewer computational resources than its classical deep learning counterparts. The SWAP-test–based fidelity objective ensures that compression preserves task-relevant structure, and the trainable quantum kernel further enhances separability in the latent space. Extensive experiments under both ideal and noisy quantum regimes confirm the robustness of the proposed framework. Notably, quantum noise, rather than degrading performance, introduces a stabilizing regularization effect, improving generalization over unseen data. Deployment on real IBM Quantum hardware further validates the method’s practicality and hardware-awareness under constrained conditions. Beyond its empirical performance, this study underscores quantum autoencoders as a promising direction for resource-efficient, adaptive security analytics in emerging quantum–classical infrastructures. Future work can explore larger qubit systems, hybrid noise mitigation, and hardware-calibrated ansatz design to scale the framework toward real-time, high-throughput quantum cybersecurity solutions.

\bibliographystyle{ieeetr}
\bibliography{ref}

@article{soe2019rule,
  title={Rule generation for signature based detection systems of cyber attacks in iot environments},
  author={Soe, Yan Naung and Feng, Yaokai and Santosa, Paulus Insap and Hartanto, Rudy and Sakurai, Kouichi},
  journal={Bulletin of Networking, Computing, Systems, and Software},
  volume={8},
  number={2},
  pages={93--97},
  year={2019}
}

@article{pang2021deep,
  title={Deep learning for anomaly detection: A review},
  author={Pang, Guansong and Shen, Chunhua and Cao, Longbing and Hengel, Anton Van Den},
  journal={ACM computing surveys (CSUR)},
  volume={54},
  number={2},
  pages={1--38},
  year={2021},
  publisher={ACM New York, NY, USA}
}

@article{hdaib2024quantum,
  title={Quantum deep learning-based anomaly detection for enhanced network security},
  author={Hdaib, Moe and Rajasegarar, Sutharshan and Pan, Lei},
  journal={Quantum Machine Intelligence},
  volume={6},
  number={1},
  pages={26},
  year={2024},
  publisher={Springer}
}

@article{sewak2020overview,
  author={Gupta, Rishabh and Saxena, Deepika and Gupta, Ishu and Makkar, Aaisha and Kumar Singh, Ashutosh},
  journal={IEEE Networking Letters}, 
  title={Quantum Machine Learning Driven Malicious User Prediction for Cloud Network Communications}, 
  year={2022},
  volume={4},
  number={4},
  pages={174-178},
  doi={10.1109/LNET.2022.3200724}}

@article{oh2024quantum,
  title={Quantum support vector data description for anomaly detection},
  author={Oh, Hyeondo and Park, Daniel K},
  journal={Machine Learning: Science and Technology},
  volume={5},
  number={3},
  pages={035052},
  year={2024},
  publisher={IOP Publishing}
}

@inproceedings{hdaib2023quantum,
  title={Quantum autoencoder frameworks for network anomaly detection},
  author={Hdaib, Moe and Rajasegarar, Sutharshan and Pan, Lei},
  booktitle={International Conference on Neural Information Processing},
  pages={69--82},
  year={2023},
  organization={Springer}
}

@article{singh2025modeling,
  title={Modeling Feature Maps for Quantum Machine Learning},
  author={Singh, Navneet and Pokhrel, Shiva Raj},
  journal={arXiv preprint arXiv:2501.08205},
  year={2025}
}

@inproceedings{mirsky2018kitsune,
  title        = {Kitsune: An Ensemble of Autoencoders for Online Network Intrusion Detection},
  author       = {Mirsky, Yisroel and Doitshman, Tomer and Elovici, Yuval and Shabtai, Asaf},
  booktitle    = {2018 Network and Distributed System Security Symposium (NDSS)},
  year         = {2018},
  organization = {Internet Society}
}

@inproceedings{javaid2016deep,
  title        = {A Deep Learning Approach for Network Intrusion Detection System},
  author       = {Javaid, Ahmad and Niyaz, Qaiser and Sun, Weiqing and Alam, Mansoor},
  booktitle    = {Proceedings of the 9th EAI International Conference on Bio-inspired Information and Communications Technologies (formerly BIONETICS)},
  pages        = {21--26},
  year         = {2016},
  publisher    = {ICST}
}

@misc{sahin2025qiskitmachinelearningopensource,
      title={Qiskit Machine Learning: an open-source library for quantum machine learning tasks at scale on quantum hardware and classical simulators}, 
      author={M. Emre Sahin and Edoardo Altamura and Oscar Wallis and Stephen P. Wood and Anton Dekusar and Declan A. Millar and Takashi Imamichi and Atsushi Matsuo and Stefano Mensa},
      year={2025},
      eprint={2505.17756},
      archivePrefix={arXiv},
      primaryClass={quant-ph},
      url={https://arxiv.org/abs/2505.17756}, 
}

@article{Chandrasekhar2025,
  author    = {Chandrasekhar, S. and Pokhrel, S. R. and Singh, N.},
  title     = {Adapting Quantum Machine Learning for Energy Dissociation of Bonds},
  journal   = {ChemRxiv},
  year      = {2025},
  doi       = {10.26434/chemrxiv-2025-9x650}
}

\end{document}